\documentclass[letterpaper, 10 pt, conference]{ieeeconf}  %

\IEEEoverridecommandlockouts                              %

\usepackage{graphicx}

\usepackage{tikz}
\usepackage{comment}
\usepackage{amsmath,amssymb} %
\usepackage{color}
\usepackage{colortbl}
\usepackage{multicol}
\usepackage{multirow}
\usepackage{cite}
\usepackage{lipsum}
\usepackage{booktabs}

\newcommand{\tikzxmark}{%
\tikz[scale=0.23] {
    \draw[line width=0.7,line cap=round] (0,0) to [bend left=6] (1,1);
    \draw[line width=0.7,line cap=round] (0.2,0.95) to [bend right=3] (0.8,0.05);
}}
\newcommand{\tikzcmark}{%
\tikz[scale=0.23] {
    \draw[line width=0.7,line cap=round] (0.25,0) to [bend left=10] (1,1);
    \draw[line width=0.8,line cap=round] (0,0.35) to [bend right=1] (0.23,0);
}}

\usepackage[accsupp]{axessibility}  %

\usepackage{url}
\usepackage{hyperref}
\newcommand{\etc}{\textit{etc} }

\newcommand{\ie}{\textit{i}.\textit{e}., }

\definecolor{Gray}{gray}{0.9}

\title{\LARGE \bf
Kalib: Easy Hand-Eye Calibration with Reference Point Tracking
}

\author{Tutian Tang$^{1*}$, Minghao Liu$^{1*}$, Wenqiang Xu$^{1}$ and Cewu Lu$^{1}$%
\thanks{*Equal contribution.}%
\thanks{$^{1}${\tt\small \{tttang, lmh209, vinjohn, lucewu\} @sjtu.edu.cn}. Tutian Tang, Minghao Liu and Wenqiang Xu are with School of Electronic Information and Electrical Engineering, Shanghai Jiao Tong University, Shanghai, China, and also with the Meta Robotics Institute, Shanghai Jiao Tong University, Shanghai, China. Cewu Lu is the corresponding author, a member of Qing Yuan Research Institute and MoE Key Lab of Artificial Intelligence, AI Institute, Shanghai Jiao Tong University, Shanghai, China.}%
}

\begin{document}

\maketitle
\thispagestyle{empty}
\pagestyle{empty}

\begin{abstract}

Hand-eye calibration aims to estimate the transformation between a camera and a robot. Traditional methods rely on fiducial markers, which require considerable manual effort and precise setup.
Recent advances in deep learning have introduced markerless techniques but come with more prerequisites, such as retraining networks for each robot, and accessing accurate mesh models for data generation.
In this paper, we propose Kalib, an automatic and easy-to-setup hand-eye calibration method that leverages the generalizability of visual foundation models to overcome these challenges.
It features only two basic prerequisites, the robot's kinematic chain and a predefined reference point on the robot.
During calibration, the reference point is tracked in the camera space. Its corresponding 3D coordinates in the robot coordinate can be inferred by forward kinematics. Then, a PnP solver directly estimates the transformation between the camera and the robot without training new networks or accessing mesh models.
Evaluations in simulated and real-world benchmarks show that Kalib achieves good accuracy with a lower manual workload compared with recent baseline methods.
We also demonstrate its application in multiple real-world settings with various robot arms and grippers.
Kalib's user-friendly design and minimal setup requirements make it a possible solution for continuous operation in unstructured environments.
The code, data, and supplementary materials are available at \url{https://sites.google.com/view/hand-eye-kalib}.

\end{abstract}

\section{Introduction}

Hand-eye calibration, which is to estimate the transformation between the camera and the robot, is one of the fundamental problems for robot manipulation with a long history~\cite{park_robot_1994,horaud_hand-eye_1995,easy-hand-eye-source,antonello_fully_2017}. A precise and easy-to-use hand-eye calibration method not only facilitates the successful interaction with the environment, but also enables researchers to rapidly iterate the setups and downstream algorithms in laboratories.

Traditional methods usually rely on calibration boards with fiducial markers such as AprilTag~\cite{apriltag} and ArUco~\cite{aruco}, which must be printed with precise dimensions, mounted flatly on a rigid board, and carefully positioned. In the industry, these well-established methods are widely used. They can achieve very high precision on condition that certain procedures are strictly followed.
In contrast, in real-world deployments such as household environments, end users are often unwilling and even unable to engage in strict and time-consuming calibration processes.
In addition, in these unstructured settings, accidents may result in recalibration for example the robot colliding with cameras or tripods. In such cases, an easier-to-setup calibration method that eliminates the need for printed fiducial markers becomes significant, as it can reduce human labor and facilitate continuous operation.

Prior research works have explored the use of neural networks to predict various semantic representations of the robot, including structural keypoints (\ie joints)~\cite{DREAM}, 6D poses~\cite{sefercik_learning_2022, lu_markerless_2023}, and masks~\cite{easyhec}.
These methods leverage high-level semantic vision models, successfully eliminating reliance on calibration boards.
However, they also introduce additional efforts, prerequisites, and constraints.
First, the models need to be trained for each robot. When the robot's appearance is altered, for example, due to the installation of extra sensors, new end-effectors, or electronic skins, the networks must be retrained.
Second, generating training data requires access to the robot's precise mesh models and textures, which may not always be available.
Third, some methods rely on high-precision depth sensors, which are not widely available. Others are designed only for either the \textit{eye-in-hand} or the \textit{eye-on-base} setting, making the use case limited. 

\begin{figure}
    \centering
    \includegraphics[width=0.9\linewidth]{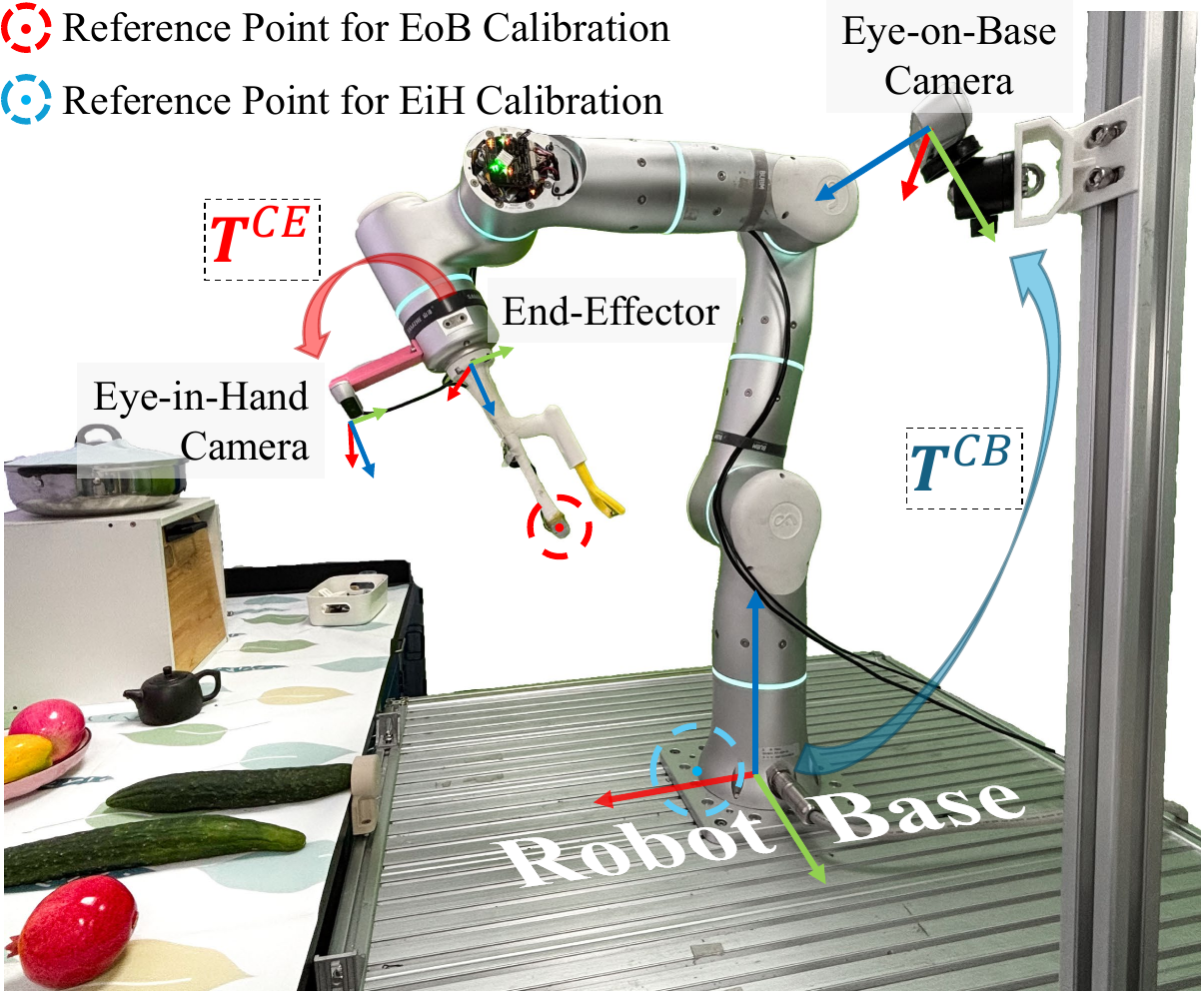}
    \caption{Hand-eye calibration estimates the transform \textbf{\textit{T}} between the camera and the robot. The proposed method can solve both eye-on-base (EoB) and eye-in-hand (EiH) calibration by tracking a predefined reference point. It's designed to work with minimal manual effort in household and unstructured environments, featuring an automatic pipeline and self-contained setup.}
    \label{fig:teaser}
\end{figure}

\begin{table*}
\setlength{\tabcolsep}{10pt}
\centering
\caption{Comparison with Existing Methods.}
\label{tab:related_work}
\begin{tabular}{c c c c c c c c}
\toprule
Category & Type & Markerless & Setting & Prerequisites & No-Training & Occlusion & Workload \\ \midrule
\multirow{2}{*}{Traditional} & Board~\cite{antonello_fully_2017} & \tikzxmark & \textbf{Both} & \textbf{Kinematics} & \tikzcmark & / & Heavy \\
 & SfM~\cite{heller_structure--motion_2011} & \tikzcmark & Eye-in-Hand & \textbf{Kinematics} & \tikzcmark & / & \textbf{Low} \\ \midrule
\multirow{4}{*}{\shortstack{Learning-\\Based}} & Keypoint~\cite{DREAM} & \tikzcmark & Eye-on-Base & + Mesh Model & \tikzxmark& / & \textbf{Low} \\
 & Mask~\cite{easyhec} & \tikzcmark & Eye-on-Base & + Mesh Model & \tikzxmark& \tikzxmark& Medium \\
 & 6D Pose~\cite{sefercik_learning_2022} & \tikzcmark & Eye-on-Base & + Mesh Model & \tikzxmark& / & \textbf{Low} \\
 & \textbf{Ours} & \tikzcmark & \textbf{Both} & \textbf{Kinematics} & \tikzcmark & \tikzcmark& \textbf{Low} \\
\bottomrule
\end{tabular}
\begin{minipage}{0.92\textwidth}
For each type, we select one representative method for comparison.
\textbf{Markerless}: We mark \tikzcmark if no fiducial marker is required.
\textbf{Setting}: Whether the method can work under eye-in-hand or eye-on-base setting, or both.
\textbf{Prerequisites}: Kinematics means the kinematic model is required, which is the basic requirement for calibration. Mesh model means the precise mesh model is also required.
\textbf{No-Training}: We mark \tikzcmark if there is no need to re-train neural networks on a new setup.
\textbf{Occlusion}: Whether the method can work under occlusion. / means not applicable or not reported.
\textbf{Workload} compares the relative amount of manual work.
\textbf{Bold items} are considered superior to others.
\end{minipage}
\end{table*}

We note that hand-eye calibration fundamentally involves establishing the correspondence between specific pixels in the image and their respective coordinates on the robot.
Our key insight is that the hand-eye calibration problem can be solved with only two basic prerequisites, \ie the robot kinematic chain and a reference point attached to it.
We thus propose \textbf{Kalib}.
In each calibration process, we define a reference point on the robot and move it naturally in the workspace.
The 3D coordinates of the reference point in the robot space are derived using forward kinematics.
An off-the-shelf foundation model tracks its 2D coordinates in the image frame, fully leveraging the temporal priors from the robot's movement.
Finally, a Perspective-n-Point (PnP) solver~\cite{epnp} estimates the camera-to-robot transformation and completes the calibration.

To evaluate our method, we first measure its accuracy in a simulation environment. The mean translational and rotational errors are around 0.3 cm and 0.5 degrees, respectively.
We then conduct extensive experiments on DROID~\cite{droid}, a large-scale, in-the-wild dataset for robot manipulation.
Results show that our method can work in household environments with noisy backgrounds. Its markerless nature enables it to serve as a remedial solution when traditional methods, used for building the dataset, produce significant errors by accident. Finally, we showcase the real-world application of the proposed method across various settings and robots, including UR10, xArm, ShadowHand, \etc.

Our main contribution of the \textbf{Kalib} method is user-friendly and easy-to-setup hand-eye calibration.
It works with common RGB cameras under both the \textit{eye-in-hand} and the \textit{eye-on-base} settings without training any networks, further reducing manual labor. The automatic pipeline and self-contained setup (\ie no fiducial tag or board required) unleash the potential for continuous operation in unstructured environments.

\section{Related Work}
Our work is most closely related to hand-eye calibration methods and point tracking with foundation models.

\subsection{Hand-Eye Calibration}
Hand-eye calibration has been studied for a long time.
There are roughly two different settings, namely, \textit{eye-on-base} and \textit{eye-in-hand}.
Traditional methods~\cite{park_robot_1994,horaud_hand-eye_1995,antonello_fully_2017,6088553,doi:10.1177/02783649922066213} rely on calibration boards with fiducial markers~\cite{apriltag,aruco,artag} or canonical objects~\cite{yang2018robotic}. These markers should be printed, mounted, and positioned with care. The 6D poses of the markers are inferred, and the camera-to-robot transformation can be solved as a linear system.
Not only does the whole process involve certain manual labor, but also many detailed factors can lead to incorrect marker detection and finally result in failure~\cite{easy-hand-eye-source}.
Therefore, the need for markerless calibration methods to save manual labor lasts long. The pioneers~\cite{heller_structure--motion_2011,andreff_robot_2001,zhi_simultaneous_2017} adopt Structure-from-Motion (SfM) techniques to estimate the camera motion during the calibration process, which only work under the eye-in-hand setting.

Later, with the advancement of deep learning, many works utilize neural networks to calibrate the camera, especially under the eye-on-base setting.
For example, DREAM~\cite{DREAM} and CtRNet~\cite{lu_markerless_2023} adopt neural networks to detect the 2D structural keypoints of the robot and use the PnP algorithm to estimate the camera pose. The idea behind this is that the robot itself is a standardized item with rich visual features, so its structural keypoints (\ie joints) can act as the marker for calibration. 
Other works adopt 6D pose estimation networks~\cite{sefercik_learning_2022,Valassakis2021LearningEC} to directly estimate the transformation of the robot arm in the camera coordinate.
However, the direct 6D pose estimation is reported to be noisy and sensitive to many factors, including light conditions, occlusion, and the sim-to-real gap~\cite{foundationPose}. Therefore, the RGB-D cameras are often used to acquire the point cloud so that the ICP algorithms~\cite{icp} can further refine the results.
A more recent work, EasyHeC~\cite{easyhec}, first estimates the 6D pose and further refines it through the mask prediction and differentiable rendering mechanism, which also requires the precise mesh model.
However, mesh quality has a direct impact on calibration results.
All these methods involve training networks for specific robots. The trained networks can not generalize to new robots, so users must retrain them on a new setup.
In comparison, the proposed \textbf{Kalib} adopts visual foundation models to prevent the users from training any networks and eliminate the need for the precise mesh model.
Moreover, \textbf{Kalib} can solve both eye-in-hand and eye-on-base settings, even under self-occlusion, making it especially suitable for today's research community that is pushing intelligent robots into unstructured environments~\cite{droid}. We summarize the comparison in Table~\ref{tab:related_work}.

\begin{figure*}
    \centering
    \includegraphics[width=0.9\linewidth]{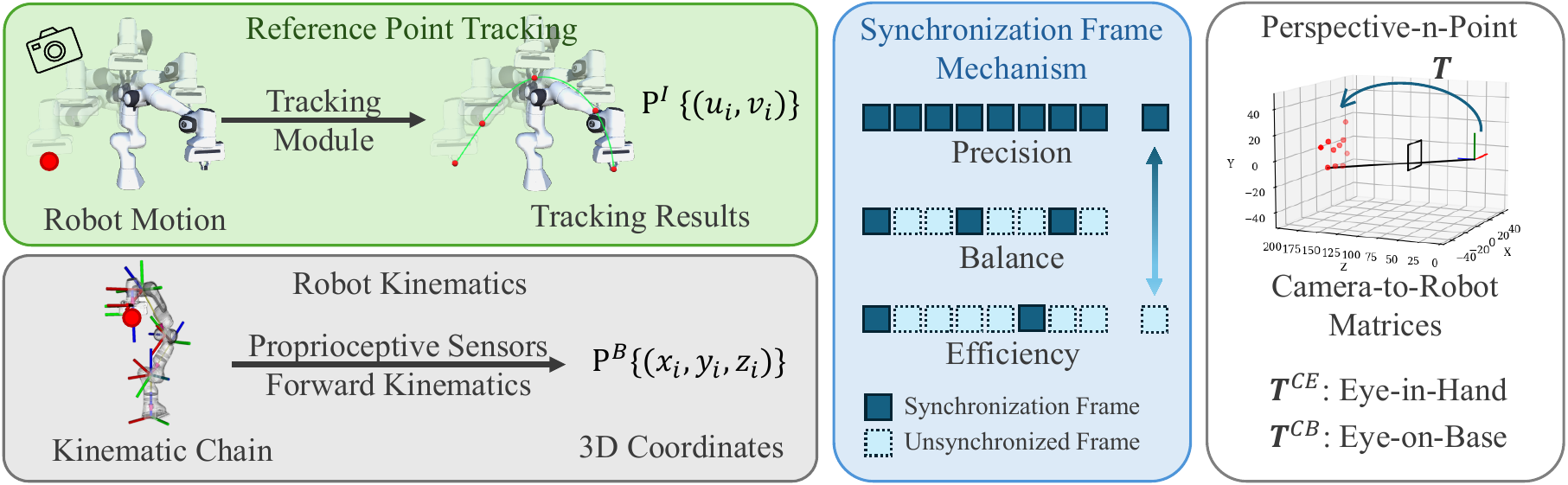}
    \caption{
    The whole pipeline starts by defining a reference point on the kinematic chain (Sec.~\ref{sec:method_tracking_target}).
    The reference point tracking module can track its 2D position in the image frame (Sec.~\ref{sec:method_point_tracking_module}), while its 3D coordinates in the robot frame can be derived by forward kinematics (Sec.~\ref{sec:method_robot_kinematics}). Here the synchronization frame mechanism is introduced to balance precision and efficiency (Sec.~\ref{sec:method_sync}). Finally, the PnP module can estimate the camera-to-robot transformation matrix, either $\mathbf{T}^{CE}$ for the eye-in-hand setting (Sec.~\ref{sec:method_pnp}) or $\mathbf{T}^{CB}$ for the eye-on-base setting (Sec.~\ref{sec:method_eih}).
    }
    \label{fig:pipeline}
\end{figure*}

\subsection{Point Tracking with Foundation Models}

Point tracking, also known as pixel tracking, involves recovering the motion of certain pixels across video frames. Early studies formulate it as an optical flow estimation problem~\cite{opticalflow_1,opticalflow_2} or particle tracking problem~\cite{particle_1,particle_survey}. However, these methods primarily focus on local image features and thus lack the ability for long-term tracking. Recent advances in visual foundation models introduce extensive pre-training on large datasets~\cite{mae,dino}, enabling the encoder networks to extract rich features and generalize across natural images, making it possible to track any pixels in long video sequences~\cite{tapir,dino-tracker}.
Co-Tracker~\cite{cotracker} can jointly track the pixels representing the same instance to improve accuracy and robustness.
SpatialTracker~\cite{SpatialTracker} lifts the 2D pixels into the 3D space, making it robust under heavy occlusion and complex motion.
MatchAnything~\cite{matchanything} further pushes the boundaries, enabling cross-modal matching and tracking.

\section{Method}
\label{sec:method}

\subsection{Overview}
Given the robot kinematics and the intrinsic camera parameters, hand-eye calibration aims to estimate the transformation between the camera and the robot system. In this paper, we formulate it into a Perspective-n-Point (PnP) problem. The camera intrinsic parameters are defined by the $3\times 3$ intrinsic matrix $\mathbf{K}$, while transformations are represented by $4\times 4$ matrices $\mathbf{T}\in SE(3)$.
Depending on the scenario, the camera is mounted either with a fixed transformation to the robot base $\mathbf{T}^{CB}$ in the \textit{eye-on-base} configuration, or with a fixed transformation to the end-effector $\mathbf{T}^{CE}$ in the \textit{eye-in-hand} configuration.
Note that these two settings are fundamentally dual, as the transformation between the base and end-effector $\mathbf{T}^{BE}$ can be derived by forward kinematics and the relationship $\mathbf{T}^{CE}=\mathbf{T}^{CB}\mathbf{T}^{BE}$ always holds.
Therefore, we will first explain the proposed method in the eye-on-base setup in Section~\ref{sec:method_2d} and Section~\ref{sec:method_pnp}, and then demonstrate how it works under the eye-in-hand setup in Section~\ref{sec:method_eih}.

Figure~\ref{fig:pipeline} shows the overall pipeline of the proposed method. 
With the tracking module, our method makes full use of the spatial-temporal information generated from the movement of the robot, thus making the calibration process automatic.
Given $N$ consecutive frames that record the movement of the robot, we first select a reference point on the robot and track its 2D position $P_i^I (u_i, v_i)$ on each image $i$. 
Simultaneously, we record the joint positions $\theta_i$ at each frame with the proprioceptive sensors integrated into the robot. 
The 3D coordinate of the reference point in the robot base coordinates $P_i^B (x_i, y_i, z_i)$ can therefore be derived by forward kinematics. 
Finally, the PnP module will estimate the camera-to-robot transformation matrix $\mathbf{T}^{CB}$.

\subsection{Temporal Point Tracking}
\label{sec:method_2d}

\subsubsection{Tracking Target}
\label{sec:method_tracking_target}
Before the tracking process begins, we need to decide the target point to track.
A potential candidate should meet the following three conditions.
First, it should be attached rigidly to the kinematic chain so that we can precisely determine its 3D coordinates on the robot with the proprioceptive sensors and forward kinematics.
Second, it should be generally unobstructed in the camera view during the robot's movement.
Third, it should correspond to clear visual features on the robot's surface so that the tracking module can work precisely.
Unlike current structural keypoint prediction methods, which predict the coordinates of the robot joints in the image, our method should track the target point on the robot's surface. Otherwise, the projected points of internal joints onto the robot's surface may vary with different movements and cause unreliable tracking results. The details will be discussed in Section~\ref{sec:choice_kpts}.

With these considerations in mind, we discuss some typical choices of reference points.
For robot arms without an end-effector, the center point on the flange for mounting the end-effector is a good choice.
For robots with parallel grippers, we can choose the center of the fingertips when the gripper is closed as the reference point.
For robots with customized tools, the tip of the tool is usually slim and clear, making it suitable as the reference point, as illustrated in Figure~\ref{fig:teaser}.
Also, the screw cap or screw hole on the end-effector flange mount can also be a good choice since the flange mount is usually a standard machining part with precise dimensions, so we can easily get its exact transformation relative to the robot kinematic tree. Please refer to the supplementary video for more details.

\subsubsection{Point Tracking Module}
\label{sec:method_point_tracking_module}
Given $N$ camera frames and the initial position of the point $P_1^I (u_1, v_1)$, the tracking module will give consecutive positions of the target on each image frame $\mathbf{P}^I = \{(u_i, v_i)|i=1,2,...,N\}$.
The initial position of the reference point on the first frame can be annotated by a single mouse click, which is much easier than annotating (or refining) the mask or 6D poses in previous methods.
Moreover, if the camera and robot's poses do not dramatically change compared with the previous calibration, we may reuse the previous annotation.
We can use visual foundation models to track the reference point on consecutive images.
To note, the off-the-shelf foundation models pre-trained on large-scale datasets already demonstrate excellent generalizability and zero-shot accuracy, so we do \textbf{not} need to retrain or finetune the networks. This is exactly the application scenario we want.
We evaluate some popular foundation models~\cite{cotracker,dino-tracker,SpatialTracker} in our use cases, with qualitative results available in the supplementary video.
We finally decided to use SpatialTracker~\cite{SpatialTracker} as the tracking module because it achieves a good balance between accuracy and inference speed, and it can robustly track the point even under noisy and challenging environments. Obviously, our method will directly benefit from the future development of more robust and accurate point-tracking models.

\subsubsection{Robot Kinematics}
\label{sec:method_robot_kinematics}
In each calibration process, the joint positions of the robot at each frame are recorded $\Theta = \{ (j_{i1}, j_{i2}, ..., j_{iJ}) | i=1,2,...,N \}$, where $N$ is the number of recorded frames, and $J$ is the number of controllable joints. We can then use forward kinematics~\cite{dh} to compute the 3D coordinates of the reference point $\mathbf{P}^B = \{(x_i, y_i, z_i)|i=1,2,...,N\}$ in the robot base coordinate.

\subsubsection{Synchronization Mechanism}
\label{sec:method_sync}
In practice, when the robot is moving, motion blur may happen and we may observe slight timing mismatches (\ie asynchronization) between the robot and the camera. We hence introduce some \textit{synchronization frames}, in which we ensure the robotic arm completely stops before the camera capture. In this way, we can get stable readings from robot proprioceptive sensors and synchronized frames from the camera.

There are various strategies to determine how often to introduce synchronization frames during the robot movement.
One approach is to make every frame a synchronization frame, where the robotic arm pauses after each small movement. While this method is time-consuming, it yields the most precise results.
For best efficiency, synchronization frames can be introduced at intervals, which significantly reduces pauses.
Those unsynchronized frames between two synchronized frames should still be retained and processed by the tracking module, to prevent the module from losing track due to excessive motion between two consecutive synchronized frames. However, the unsynchronized tracking results should not be used for later camera pose estimation.
Experimental results in Section~\ref{sec:num_of_kps} indicate that as few as 10-20 synchronization frames can be enough for a reasonable calibration result, while it gradually improves with more frames.
In Section~\ref{sec:exp_tracking}, we will demonstrate that the tracking algorithm can reliably track over hundreds of frames, a number sufficient to achieve accurate calibration.
To note, our method requires manual annotation only for the first frame. The subsequent process is entirely automatic and requires no additional effort. This flexibility allows users to balance efficiency and precision according to their specific use case.

\subsection{Camera Pose Estimation}
\label{sec:method_pnp}

We formulate the camera pose estimation problem as a Perspective-n-Point (PnP) problem.
Following the perspective projection model for cameras, given $\mathbf{K}$, $\mathbf{P}^I$ and $\mathbf{P}^B$, each 2D point $P_i^I(u_i, v_i) \in \mathbf{P}^I$ in image frame $i$ and the corresponding 3D coordinates in the robot base coordinate $P_i^B(x_i, y_i, z_i) \in \mathbf{P}^B$ obey the following equation:
\begin{equation}
\label{eqn:pnp_eob}
s\cdot \hat{P}_i^I
=
\hat{\mathbf{K}}
\cdot
\mathbf{T}^{CB}
\cdot
\hat{P}_i^B ,
\end{equation}
where $s$ is a scaling factor, $\mathbf{T}^{CB}$ is the desired camera extrinsic matrix, 
$\hat{\mathbf{K}}$ is the homogeneous camera intrinsics, and $\hat{P}$ is the homogeneous form of $P$.
With a minimum of three pairs of corresponding points, a PnP solver can solve $\mathbf{T}^{CB}$ in the above linear system.
We adopt the SQPnP solver~\cite{sqpnp} implemented in OpenCV~\cite{opencv_library}, as it can deliver consistently fast and global optimal solutions.

Unlike those structural keypoint-based methods which predict multiple keypoints on a single frame and treat the robot's motion as several discrete single images, our method can work with as few as only one pair of points per frame. It leverages spatial-temporal information across frames to produce sufficient point pairs for the PnP solver. By densely tracking the reference point over time, our method can mitigate the impact of noise on the PnP algorithm~\cite{sqpnp}, as will be discussed in Section~\ref{sec:num_of_kps}.
In addition, to avoid some known limitations of PnP, we should avoid the points being arranged collinearly~\cite{aruco}. This means the reference point should ideally traverse the workspace under the camera's view as much as possible instead of being confined to a single line.

\subsection{Eye-in-Hand Setup}
\label{sec:method_eih}
Recap that in the eye-in-hand configuration, the camera is rigidly mounted to the end-effector with the unknown transformation $\mathbf{T}^{CE}$. To simplify the problem formulation, we conceptually \textit{reverse} the topology of the robot, treating the original end-effector as the base and vice versa. This transforms the eye-in-hand setup into its duality, eye-on-base setup. Then, the reference point should naturally be on the robot base. As is illustrated in Figure~\ref{fig:teaser}, we select the intersection point of the shell of the robot base and the $x$ axis of the robot coordinate (which is usually interpreted as the \textit{forward} direction of the robot) as the reference point. Its coordinate in the robot base frame is given by $P^B_{ref} (x_{ref}, 0, 0)$, which can be measured directly and satisfies the three conditions aforementioned in Section~\ref{sec:method_2d}.

To this end, following the formulation in the previous section, we derive:
\begin{equation}
\label{eqn:pnp_eih_1}
s\cdot \hat{P}^I
=
\hat{\mathbf{K}}
\cdot
\mathbf{T}^{CE}
\cdot
\hat{P}^E ,
\end{equation}
where
$\hat{P}^E$ is the homogeneous coordinate of the reference point in the end-effector's local frame.

During the calibration process, we first manually point the camera to the base and start moving the end-effector, keeping the reference point visible in the camera frame.
With forward kinematics, we obtain the transformation from the robot base to the end-effector $\mathbf{T}^{EB}_i$ at each timestamp $i$. Then we have $\hat{P}^E_i=\mathbf{T}^{EB}_i\hat{P}_{ref}^B$.
Substituting this into Equation~\ref{eqn:pnp_eih_1}, we derive
\begin{equation}
s\cdot \hat{P}^I
=
\hat{\mathbf{K}}
\cdot
\mathbf{T}^{CE}
\cdot
\mathbf{T}^{EB}_i\hat{P}_{ref}^B.
\end{equation}
Finally, we can again use PnP to solve $\mathbf{T}^{CE}$.

\section{Experiment}

\begin{figure*}
    \centering
    \includegraphics[width=0.9\linewidth]{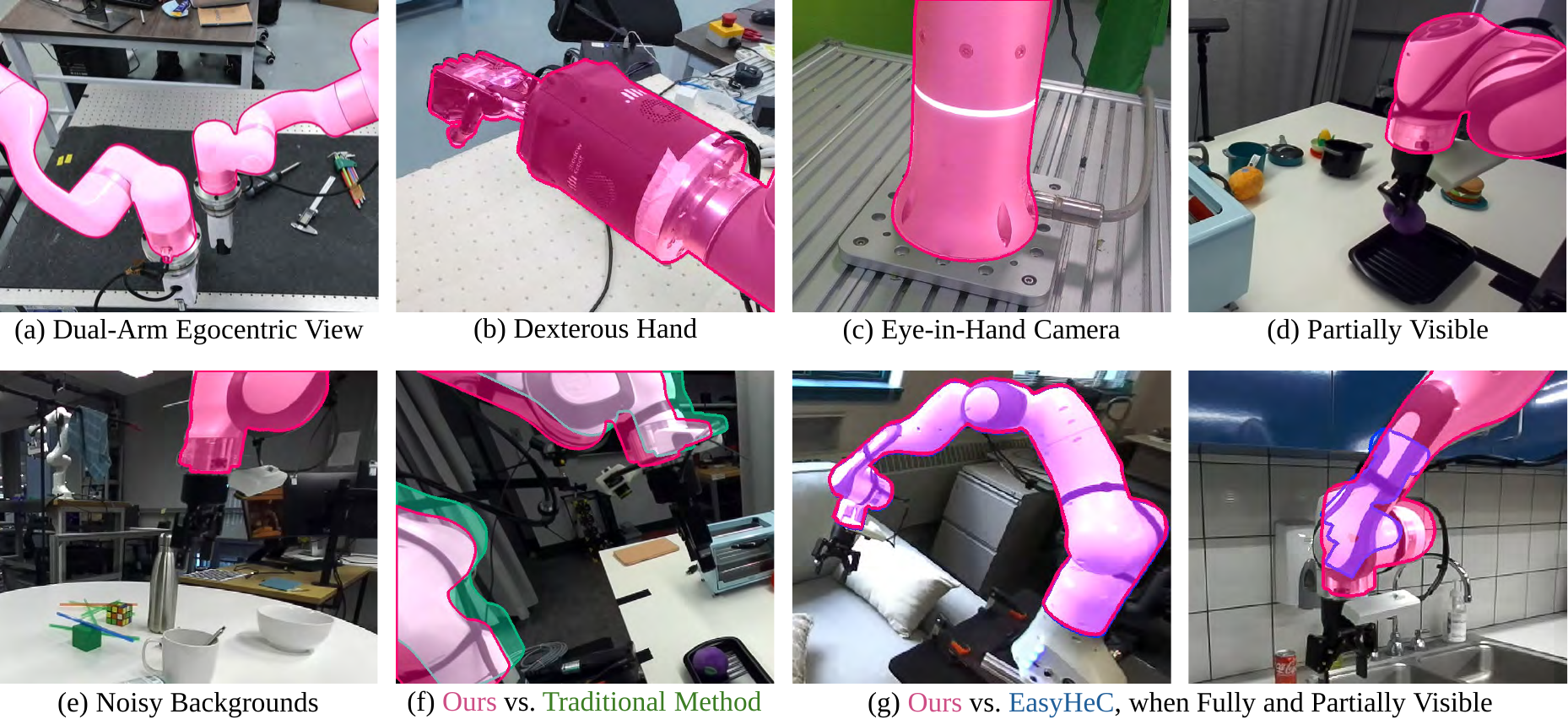}
    \caption{Qualitative results in the real world. 
    We draw masks of the robot projected onto the camera frame with our method in \textcolor{magenta}{red}.
    The precise fits of the mask and the robot suggest an accurate calibration result.
    Our method works under various settings, for example, \textbf{(a)} with a dual-arm robot and an egocentric camera, \textbf{(b)} with dexterous hand, \textbf{(c)} under the eye-in-hand setting, \textbf{(d)} when the robot is only partially visible, and \textbf{(e)} when the background is noisy.
    \textbf{(f): }
    When traditional methods fail, indicated by the \textcolor{green}{green} masks, our method can work as a post-doc remedy thanks to its markerless nature.
    \textbf{(g): }
    EasyHec (in \textcolor{blue}{blue} masks) works well with a full view of the robot but may fail with a partial view. Our method can work in both conditions.
    }
    \label{fig:qualitative}
\end{figure*}

We evaluate the proposed method in both simulated and real-world dataset. The accuracy of point tracking and hand-eye calibration are analyzed. We compare our method with DREAM~\cite{DREAM}, EasyHeC~\cite{easyhec}, and the widely-used traditional marker-based method~\cite{easy-hand-eye-source}, which is implemented in the \texttt{easy\_handeye} toolbox~\cite{easy-hand-eye-github} from the Robot Operating System (ROS)~\cite{ros}. We also demonstrate the application in various robot systems under different settings.

\subsection{Experimental Setup}
\subsubsection{Evaluation in the Simulation Environment}
We build the environment with the RFUniverse~\cite{RFU}. The Franka Emika Panda robot is placed on the ground. In the eye-on-base (EoB) setting, a virtual camera is randomly placed and pointed at the robot. We record 20 video segments, each lasting 10 seconds. Within each segment, the robot's end-effector moves towards one random direction for 1s and then switches to another, repeating this cycle 10 times.
In the eye-in-hand (EiH) setting, the camera is attached to the end-effector.
We configure the camera with a resolution of 1920 $\times$ 1080 at 30 Hz and a field of view (FOV) of 60 degrees, aligning with real-world cameras in general.
We compare the tracking and calibration results with the ground truth.

\subsubsection{Evaluation on the Real-world Dataset}
To evaluate the proposed method with real-world data, we resort to the DROID~\cite{droid} dataset, a large, diverse, high-quality dataset for robot manipulation in the wild. It contains 1.7 TB of data from 564 different scenes, in each of which a Franka robot is mounted on a mobile platform with two EoB cameras at 1280 $\times$ 720 resolution. The dataset provides RGB images and camera calibration matrices resulting from the traditional checkerboard-based calibration~\cite{opencv_library}. The backgrounds in the dataset are noisy and changeable, therefore making it challenging for tracking and calibration tasks.
Note that since the dataset is too large, we select a subset of $60$ video segments as the test set. The video list is made public.

Since there is no direct ground truth for camera calibration available in the real world, we use the reprojection error as an indirect indicator of the accuracy following previous works~\cite{easyhec,lu_markerless_2023}. To be exact, we render the masks of the Franka robot onto the camera frames based on the camera extrinsic produced by different methods and calculate the Intersection over Union (IoU) of the projected masks with the manually labeled ground truth masks on the first frame of each video. A higher IoU suggests a lower reprojection error and better accuracy.

\subsection{Results on Point Tracking}
\label{sec:exp_tracking}
\begin{figure}
    \centering
    \includegraphics[width=0.9\linewidth]{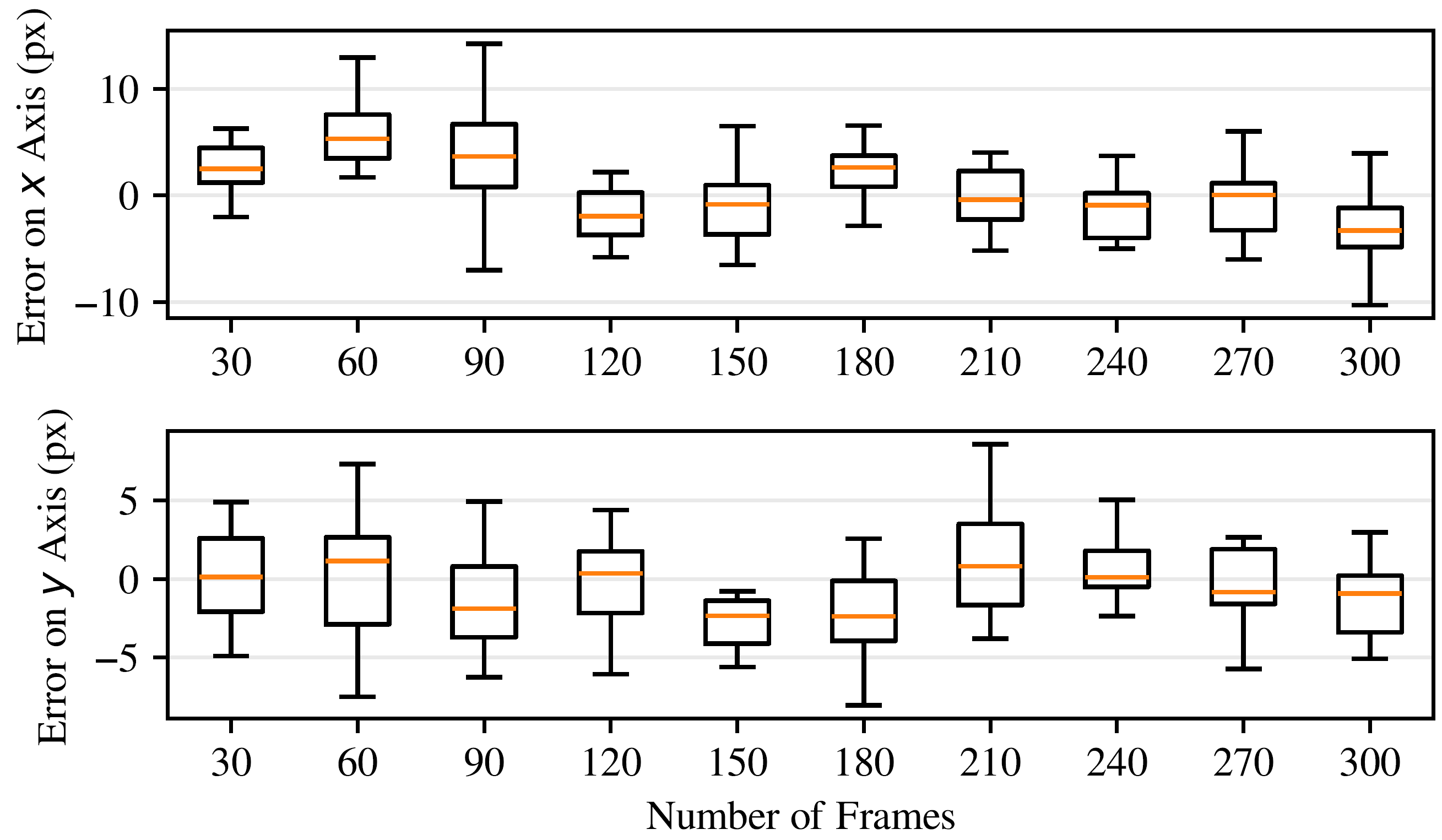}
    \caption{Error of tracking over the number of frames in simulation.}
    \label{fig:2d_track_syn}
\end{figure}

We first evaluate the zero-shot ability of the tracking module in the simulation environment.
Given the 2D position of the reference point on the first frame, we measure the error between the tracking results and the ground truth. In Figure~\ref{fig:2d_track_syn}, we show the box plot of the error on $x$ and $y$ axes over the number of frames.
The error quickly converges within $\pm 10$ pixels. It shows no sign of divergence over the whole 300 frames. A sensitivity analysis will be conducted in Section~\ref{sec:sensitivity}, which proves the $10$-pixel error is within the tolerance of the PnP algorithm. This indicates the tracking module powered by visual foundation models fully qualifies for calibration in this setting. Experiments on the real-world DROID dataset show similar results, which are shown in the supplementary video.

\subsection{Results on Hand-Eye Calibration}

\subsubsection{Evaluation in simulation environment}

\begin{table}
    \centering
    \caption{Calibration Results in simulation}
    \label{tab:exp_syn}
    \begin{tabular}{cccccc}
    \toprule
         Setting & Method&  $e_x$ (cm)&  $e_y$ (cm)&  $e_z$ (cm)&$e_r$ (rad)\\ \midrule
         \multirow{2}{*}{EiH}
         &Traditional&  0.65 & \textbf{0.42} &  1.44 & 0.08 \\
         &\textbf{Ours}&  \textbf{0.48} &  0.52 &  \textbf{0.77} & \textbf{0.07} \\
         \midrule
         \multirow{4}{*}{EoB}&Traditional&  0.68 &  1.24 &  \textbf{0.10} & \textbf{0.01}  \\
         &DREAM&  0.90   &  1.01  &  1.13  & \textbf{0.01} \\
         &EasyHeC&  0.63 &  \textbf{0.20} &  0.17 & 0.08 \\
         &\textbf{Ours}&  \textbf{0.30}&  0.45&  0.60&\textbf{0.01}\\
     \bottomrule
    \end{tabular}
    \begin{minipage}{0.95\linewidth}
    EiH stands for eye-in-hand, and EoB stands for eye-on-base. $e_x$, $e_y$, and $e_z$ represent the mean absolute translational error on $x$, $y$, and $z$ axis. $e_r$ represents the mean rotational error. 
    \end{minipage}
\end{table}

We test our hand-eye calibration pipeline in the simulation environment. In Table~\ref{tab:exp_syn}, we report the results of our method and the traditional method in both eye-in-hand (EiH) and eye-on-base (EoB) settings. We compare EasyHeC~\cite{easyhec} and DREAM~\cite{DREAM} only in the EoB setting because they are not designed for the EiH.
EasyHeC may struggle when the robot is self-occluded or partially visible within the frame.
This limitation likely arises from its differentiable renderer and gradient-based optimizer, as the masks of robot segments outside the frame do not produce gradients for optimizing the camera pose.
In comparison, the proposed method performs stably as long as the reference point is visible.

In terms of time efficiency, the tracking module can run at approximately $5$ frames per second (FPS) with a single RTX 3090 GPU, making each calibration process take around $2$ minutes. EasyHeC takes around $15$ minutes for each calibration due to the relatively slow optimization speed of its differentiable renderer module. Additionally, both EasyHeC and DREAM require several extra hours to generate the Panda robot dataset for training the networks, while the proposed method works out of the box. 

\subsubsection{Evaluation on Real-World Dataset}
On our test set, the proposed method achieves an mean IoU of $0.80$, while EasyHeC achieves $0.77$ and the traditional method used for constructing the dataset achieves $0.87$.
The slightly lower performance of our method can be attributed to the cases in the test set where the robot arm moves primarily along a straight line. In such scenarios, the PnP algorithm becomes more sensitive to noise, making it difficult to derive an accurate solution. Additionally, there are also some cases where the reference point moves out of the frame, or is occluded by the manipulated objects, which causes the tracking module to lose track. When these cases are excluded (9 out of 60 sequences), the IoU of our method improves to $0.85$.

Figure~\ref{fig:qualitative} shows a case of bad calibration from the dataset, indicating that the traditional methods can accidentally fail in real-world applications, even in expert-made datasets.
In this case, the markerless methods can be applied as a post-hoc remedy to minimize the impact of the accidents.
Again, EasyHeC works well when the robot arm is fully visible, but fails when the arm is partially visible. In comparison, the proposed method can work in both conditions.

\subsection{Ablation Studies}

\subsubsection{Number of Frames}
\label{sec:num_of_kps}

\begin{figure}
    \centering
    \includegraphics[width=0.9\linewidth]{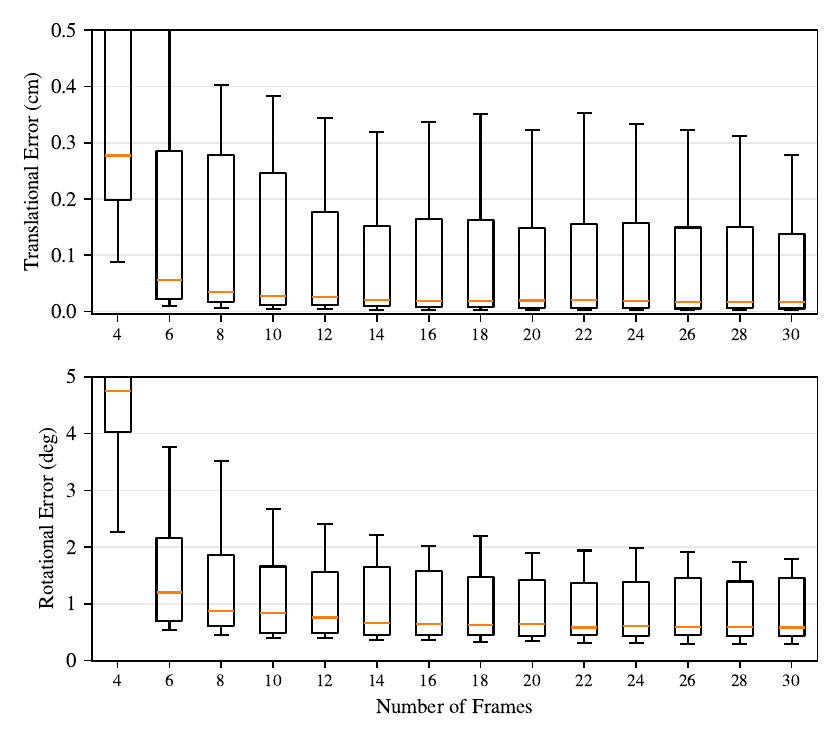}
    \caption{Translational and rotational error over the number of frames.}
    \label{fig:abl_num_frames}
\end{figure}

We analyze the influence of the number of synchronized frames (\ie pairs of corresponding points over time) on the calibration results in simulation. Since each frame can be considered ideally synchronized in the simulator, we can simply sample different numbers of corresponding points from a long sequence. Estimated camera poses from sampled points can be compared with the ground truth. We repeat the experiment on 10 random scenes.
Figure~\ref{fig:abl_num_frames} shows the mean error over number of frames. The $x$ axis starts with 4 because with 3 pairs of points, although theoretically feasible, the PnP module fails to give reasonable results. The mean error quickly converges below $0.1$ cm or $1$ degree after $10$ frames, and gradually improves with more frames. Since PnP algorithms are known to be sensitive to noise, more frames are always welcomed.

\subsubsection{Structural Keypoints vs. Reference Point}
\label{sec:choice_kpts}
\begin{figure}
    \centering
    \includegraphics[width=0.9\linewidth]{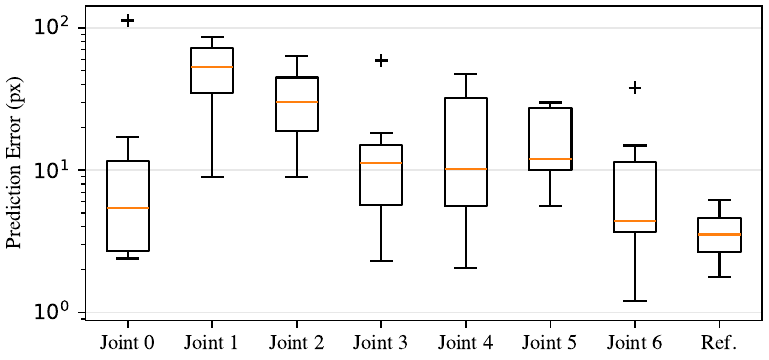}
    \caption{Prediction error in Euclidean distance of different candidates for tracking, among which the reference point achieves the lowest and the most stable tracking error. Note that the $y$ axis is on a log scale.}
    \label{fig:abl_kp_choice}
\end{figure}

Recall that our method tracks the reference point on the shell, instead of those internal structural keypoints (\ie joints), since the projected points of internal joints onto the robot's surface may vary with different movements, thus causing unreliable tracking results. To verify it, we try to use the tracking module to track those structural keypoints, namely, Joint 0 to Joint 6, over the 300 frames in the simulation environment, and compare the tracking error with the reference point on the surface. As shown in Figure~\ref{fig:abl_kp_choice}, the reference point indeed achieves the lowest and the most stable tracking error, which confirms that we should stick with the reference point over the structural keypoints in our method.

\subsubsection{Sensitivity Analysis}
\label{sec:sensitivity}

\begin{figure}
    \centering
    \includegraphics[width=0.95\linewidth]{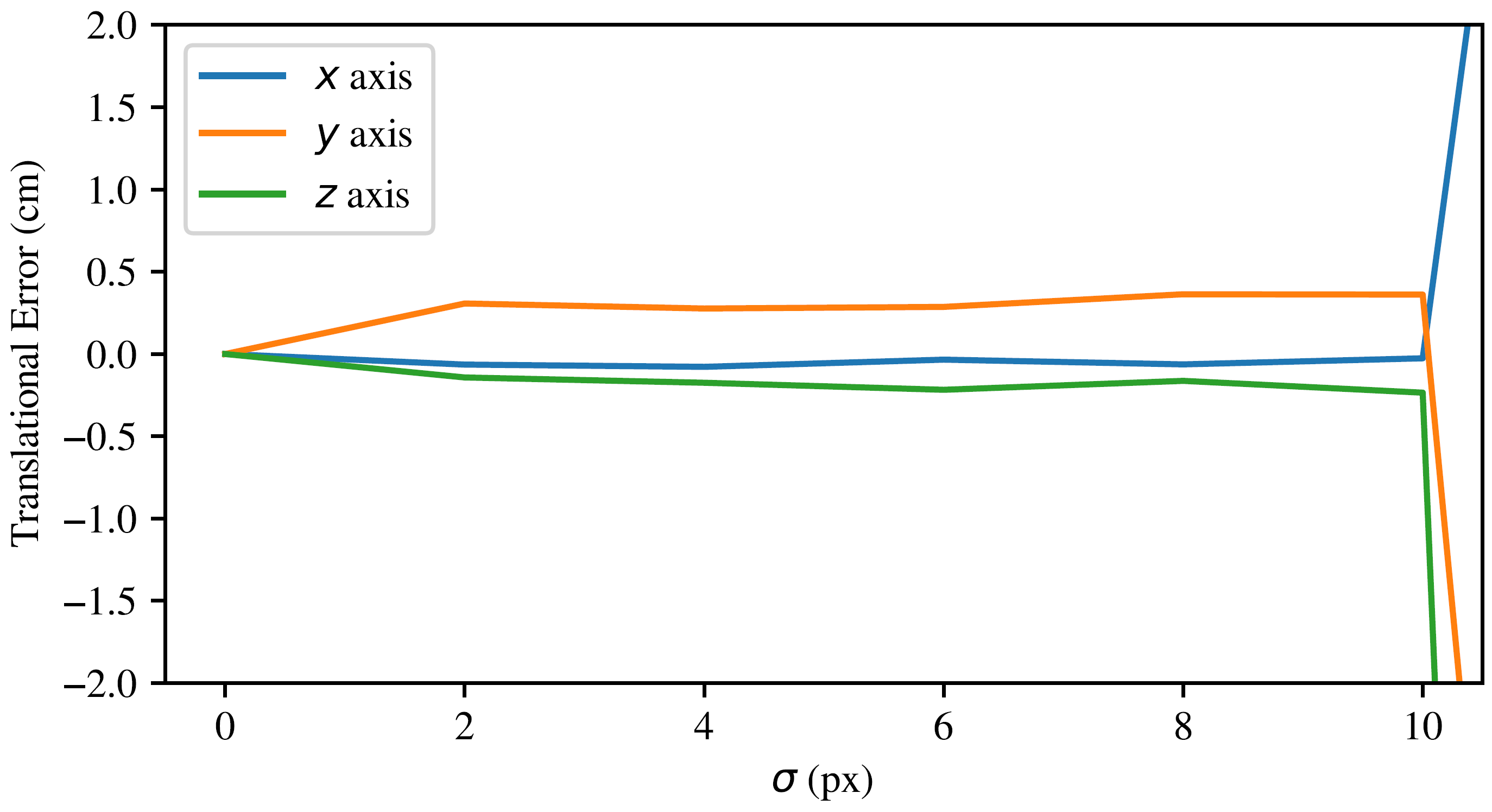}
    \caption{Sensitivity analysis. The translational error increases with a larger noise.}
    \label{fig:sensitivity-pnp}
\end{figure}

We analyze the sensitivity of the PnP module.
In the simulation environment, we first derive ground truth 2D positions $\mathbf{P}^I_{gt}$ and 3D coordinates $\mathbf{P}^B_{gt}$ of the reference point.
We then add random Gaussian noise $G(\mu, \sigma)$ to $\mathbf{P}^I_{gt}$, deriving $\mathbf{P}^{I\prime}$. The resulting $\mathbf{T}^\prime$ will be impacted by the noise. We set $\mu=0$, and $\sigma=2, 4, 6, ...$ to see how the accuracy is impacted by the noise. In Figure~\ref{fig:sensitivity-pnp}, the mean error is below $1$ cm when $\sigma \leq 10$ px. Considering the mean distance error of tracking stays within several pixels in Figure~\ref{fig:2d_track_syn}, we can conclude that the PnP algorithm can handle the noise from the tracking module in our setting.

\subsection{Application in the Real World}
We further validate our method by applying it to various robotic systems in real-world scenarios. In addition to the Franka robot, we conduct experiments on the following robotic setups: 
(1) Flexiv Rizon 4 arm: A 7-DoF robotic arm deployed in a kitchen for fruit-peeling tasks, equipped with a custom end-effector. The setup includes a third-person EoB camera and an EiH camera, both of which are Realsense D415. We demonstrate the use of Kalib to simultaneously calibrate \textbf{both} the EiH and EoB cameras.  
(2) Dual-arm mobile robot: A mobile robot base featuring two xArm 6 robotic arms, an egocentric EoB Femto Bolt camera, and two 6-DoF force sensors.  
(3) Dexterous hand: A Shadow Dexterous Hand mounted on a UR10 robotic arm, with a third-person EoB Kinect Azure camera and a specialized tactile glove worn on the hand. 
It is worth noting that these robotic systems have undergone various modifications compared with factory settings, which would typically require retraining for current neural network-based methods. However, our approach eliminates this need.
Figure~\ref{fig:qualitative} presents qualitative results, demonstrating the effectiveness of our method across these diverse scenarios.
In addition, we conduct a high-precision pointing experiment with the dual-arm robot, where a pointer attached to the gripper is directed toward the corners of a ChArUco board~\cite{aruco,opencv_library}. Accurate alignment of the pointer with the board's corners indicates low systemic error within the robotic system.
Please refer to the supplementary video for more details and qualitative results.

\section{Conclusions, Limitations, and Discussion}
In this paper, we present \textbf{Kalib}, an automatic markerless hand-eye calibration pipeline leveraging the visual foundation models for point tracking. Kalib boasts its adaptability to new setups without training any neural networks. Also, there is no need for a calibration board or the robot's exact 3D mesh model. Results show that the proposed method works well under different settings with various robot systems and highlights its potential to be integrated into real-world, unstructured scenarios.

However, the proposed method still suffers from defects in common vision-based systems. For example, the tracking module may lose track when the background is extremely noisy or motion blur occurs due to low ambient light.
Future work can focus on improving the precision and generalizability of the whole pipeline. For example, the proposed method can immediately benefit from the future advancement of tracking algorithms to perform better under the aforementioned tricky conditions.

\bibliographystyle{IEEEtran}
\bibliography{references}

\end{document}